%% file: SyntheticDataGeneration.tex
\documentclass[conference]{IEEEtran}
\IEEEoverridecommandlockouts

\usepackage{amsfonts}
\usepackage{cite}
\usepackage{amsmath}
\usepackage{algorithmic}
\usepackage{graphicx}
\usepackage{textcomp}
\usepackage[dvipsnames,table,xcdraw]{xcolor}
\usepackage{siunitx}
\usepackage{tabularx,booktabs}
\usepackage{multirow}
\usepackage{multicol}
\usepackage{caption}
\usepackage{subcaption}
\usepackage{cleveref}
\usepackage{balance}
\usepackage{lastpage}
\usepackage{tabulary}
\usepackage{breqn}
\usepackage{bbm}
\usepackage{lipsum}
\usepackage{algorithm}  
\usepackage{fancyhdr} % 添加页眉页脚
\newcolumntype{C}{>{\centering\arraybackslash}X} % centered version of "X" type
\newcolumntype{L}{>{\raggedright\arraybackslash}X}

\def\BibTeX{{\rm B\kern-.05em{\sc i\kern-.025em b}\kern-.08em
    T\kern-.1667em\lower.7ex\hbox{E}\kern-.125emX}}
    
\graphicspath{{images/}}

\begin{document}
\title{Variational Autoencoder Generative Adversarial Network for Synthetic Data Generation in Smart Home\\
}

\author{Mina Razghandi$^\dag$, Hao Zhou$^\ddag$, Melike Erol-Kantarci$^\ddag$, and Damla Turgut$^\dag$\\
{$^\dag$Department of Computer Science, University of Central Florida}\\
{$^\ddag$School of  Electrical Engineering and Computer Science, University of Ottawa }\\
mrazghandi@knights.ucf.edu, turgut@cs.ucf.edu, \{hzhou098, melike.erolkantarci\}@uottawa.ca \\
\vspace{-6 mm}
}

% packing more
\renewcommand\floatpagefraction{.9}
\renewcommand\topfraction{.9}
\renewcommand\bottomfraction{.9}
\renewcommand\textfraction{.1}
\setcounter{totalnumber}{50}
\setcounter{topnumber}{50}
\setcounter{bottomnumber}{50}
\definecolor{cadmiumgreen}{rgb}{0.0, 0.42, 0.24}
\maketitle

\thispagestyle{fancy} % 
      \lhead{} % 页眉左，需要东西的话就在{}内添加
      \chead{Accepted by 2022 IEEE International Conference on Communications (ICC) , \copyright2022 IEEE } % 页眉中
      \rhead{} % 页眉右
      \lfoot{} % 页眉左
      \cfoot{\thepage} % 页眉中
      \rfoot{} %页眉右，\thepage 表示当前页码
      \renewcommand{\headrulewidth}{0pt} %改为0pt即可去掉页眉下面的横线
      \renewcommand{\footrulewidth}{0pt} %改为0pt即可去掉页脚上面的横线
\pagestyle{fancy}

\begin{abstract}
 %Sufficient and efficient historical data is required to provide personalized experiences to smart grid customers. Data collection from residents on a daily basis is a difficult task since data must be obtained over a lengthy period of time. Assuming that environmental conditions or smart meter malfunctions have no impact on data quality, residents' privacy may be compromised. All of the aforementioned factors contribute to a scarcity of high-quality data in large volumes. A potential solution to this problem appears to be generative models. Previous research has used statistical and deep learning-based approaches to construct daily load profiles based on statistical data distribution assumptions.
%\notered{Hi All, I finished my reading. Made some corrections and left a few comments for Mina. }
Data is the fuel of data science and machine learning techniques for smart grid applications, similar to many other fields. However, the availability of data can be an issue due to privacy concerns, data size, data quality, and so on. To this end, in this paper, we propose a \textbf{Variational AutoEncoder Generative Adversarial Network (VAE-GAN)} as a smart grid data generative model which is capable of learning various types of data distributions and generating plausible samples from the same distribution without performing any prior analysis on the data before the training phase. We compared the Kullback–Leibler (KL) divergence, maximum mean discrepancy (MMD), and Wasserstein distance between the synthetic data (electrical load and PV production) distribution generated by the proposed model, vanilla GAN network, and the real data distribution, to evaluate the performance of our model. Furthermore, we used five key statistical parameters to describe the smart grid data distribution and compared them between synthetic data generated by both models and real data. Experiments indicate that the proposed synthetic data generative model outperforms the vanilla GAN network. The distribution of VAE-GAN synthetic data is the most comparable to that of real data.
\end{abstract}


\input{1_Introduction}
\input{2_RelatedWork}
\input{3_ProposedModel}
\input{4_Experiments}
\input{5_Conclusion}
\input{6_Acknowledgement}

\end{document}

%% file: 1_Introduction.tex
\section{Introduction}
\label{Introduction}
As an important part of the smart grid, smart home is expected to improve household energy usage efficiency, reduce energy cost, and enhance the user comfort level\cite{b1}. The widespread smart meter devices are considered as a key enabler of smart homes by collecting the critical data of household devices such as energy consumption profiles. These data can be further used by the smart home controllers or utility companies for load consumption and generation forecasting, demand-side management, and economic dispatch \cite{b2}. Consequently, the availability of fine-grained data becomes the prerequisite of building up a smart home.

However, access to real-world data is a challenging issue due to privacy concerns. Furthermore, the size and the quality of real-world data can also be bottlenecks for applying data science techniques to the smart grid\cite{b4}. To this end, generating synthetic data emerges as a promising alternative. Then, the generated data can be leveraged by machine learning algorithms, for instance, to decide when to implement demand response, when to charge an EV, etc. \cite{b33}. 
The existing works around synthetic data generation for smart homes can be divided into two approaches: model-based and data-driven methods. The model-based method uses mathematical equations to describe the feature of household devices, including Markov chain\cite{b5, b6}, statistical model \cite{b8}, and physical simulator-based method \cite{b7}. This approach requires extensive knowledge to build a dedicated generation model, which lacks flexibility and generalization capability. Moreover, a model-based method has difficulty in capturing the effect of user habits, which has a great impact on household power consumption. In contrast, the data-driven approaches require no prior knowledge and assumptions for the device's operation and energy consumption. It avoids the complexity of building a dedicated physical operation model of household devices and consequently increases flexibility by eliminating tedious assumptions. Recently flourishing machine learning techniques provide useful tools for synthetic data generation where the Generative Adversarial Networks (GAN) is one of the promising data generation solutions \cite{b10}. For instance, \cite{b9} proposed a deep GAN-based method to generate synthetic data for energy consumption and generation. The main idea behind the GAN network is to use a discriminator to indirectly train the generator network to produce synthetic data. The generator must deceive the discriminator in order to not distinguish between fake and real samples to reach an equilibrium point.    

In this work, we propose a novel Variational Autoencoder GAN (VAE-GAN) technique for the synthetic time series data generation of smart homes. Different from the aforementioned schemes, this approach is capable of learning various types of data distributions in a smart home and generating plausible samples from the same distribution without performing any prior analysis on the data before the training phase. In addition, utilizing a variational autoencoder network in the GAN generator module helps the network to avoid mode collapse, which is a common failure in GAN networks. More specifically, it will prevent the generator from finding only one output that seems most plausible to the discriminator and generates it every time. The main contributions of this paper are as follows:
\begin{itemize}
    \item A variational autoencoder GAN-based scheme is proposed to generate different types of time series synthetic data for smart homes with high temporal resolution.
    \item The performance of the proposed model and effectiveness of synthetically generated data are evaluated by another deep learning-based generative model, GAN, using various statistical metrics.  
\end{itemize}

The rest of this paper is organized as follows. Section~\ref{RelatedWork} introduces related work. Section ~\ref{ProposedModel} shows the proposed smart home synthetic data generation model. Section~\ref{Experiments} presents the simulation settings and results, and Section~\ref{Conclusion} concludes the paper.

%% file: 2_RelatedWork.tex
\section{Related Work}
\label{RelatedWork}
A wide variety of methods have been developed for synthetic data generation of smart grid applications. For example, a Markov chain-based user behavior simulation method is proposed in \cite{b5} for home energy consumption modeling.  \cite{b6} develops a bottom-up analysis method for the residential building energy consumption. A statistical synthetic data generator is defined in \cite{b8} for electric vehicle load modeling, in which the Gaussian
mixture model is used to estimate the connection time. In addition, \cite{b7} defines a smart residential load simulator based on MATLAB-Simulink, and it includes dedicated physical models of various household devices.

On the other hand, considering the high complexity of the above mentioned model-based methods, the data-driven methods become a favorable replacement as they do not require prior knowledge. \cite{b11} proposes a GAN-based scheme to generate synthetic labeled load patterns and usage habits, which requires no model assumptions. Furthermore, \cite{b12} introduces a model-free method for scenario generation of smart grid, and GAN is used to capture the spatial and temporal correlations of renewable power plants. Similarly, GAN is deployed in \cite{b13} to generate realistic energy consumption data by learning from actual data.

In our former work, we proposed a sequence-to-sequence learning-based method for load prediction in \cite{b14}, and a Q-learning based scheme for smart home energy management in \cite{b15}. We used a limited real dataset for our algorithm training in these former works. In this paper, we apply a novel VAE-GAN method for the synthetic data generation of the smart home. This is different from GAN-based approaches as a variational autoencoder network is deployed in the GAN generator, to overcome the mode collapse issue of the traditional GAN.

%% file: 3_ProposedModel.tex
\section{Smart Home Synthetic Data Generation}
\label{ProposedModel}
To generate realistic smart home data, we adopt a Variational Autoencoder-Generative Adversarial Network (VAE-GAN) as a data-driven approach. VAE-GAN is used to generate daily overall electricity consumption and PV production data. 

Deep learning-based generative models, which use unsupervised learning to learn data distribution and underlying patterns, have gotten a lot of attention in recent years. %Smart grid data, such as energy consumption, has greater complexity rather than repetitive seasonal data because it is tightly intertwined with residents' consumption behavior or environmental factors. Neural networks are powerful tools to overcome this complexity. 
GAN and Variational Autoencoders are two of the well-known deep learning based generative models. In the following sections, we explain the details of these techniques. 
\subsection{Generative Adversarial Network (GAN)}
\label{GANdescription}

GAN networks employ an unsupervised learning method to detect and learn patterns in input data and produce new samples that have the same distribution as the original dataset. GAN is composed of two main modules, namely generator, and discriminator, and it actively seeks an equilibrium between the two modules.
\begin{itemize}
    \item \textbf{Generator ($G$):} It maps a prior probability distribution, that is defined on input noise $p_{z}(Z)$, to a data space $G(z;\theta_{g})$, where $z$ is the input noise, and $\theta_{g}$ is the network parameter.
    \item \textbf{Discriminator ($D$):} $D(x;\theta_{d})$ produces a single scalar indicating the probability of $x$ being a member of the original data.
\end{itemize}

$G$ and $D$ play an adversarial game shown by equation (\ref{eq:minmax}), where $D$ maximizes the probability of assigning true labels, $logD(x)$, and $G$ tries to minimize the same probability: 
 \begin{dmath}
    \underset{G}{min}~\underset{D}{max}   L_{GAN}(D,G) = E_{x}[log(D(x))] + E_{z}[1-log(D(G(z)))]
    \label{eq:minmax}
\end{dmath}
%\notered{make sure to explain all notation in the text}
\begin{figure}
    \centering
        \centering
        \captionsetup{justification=centering}
        \includegraphics[width=0.5\textwidth]{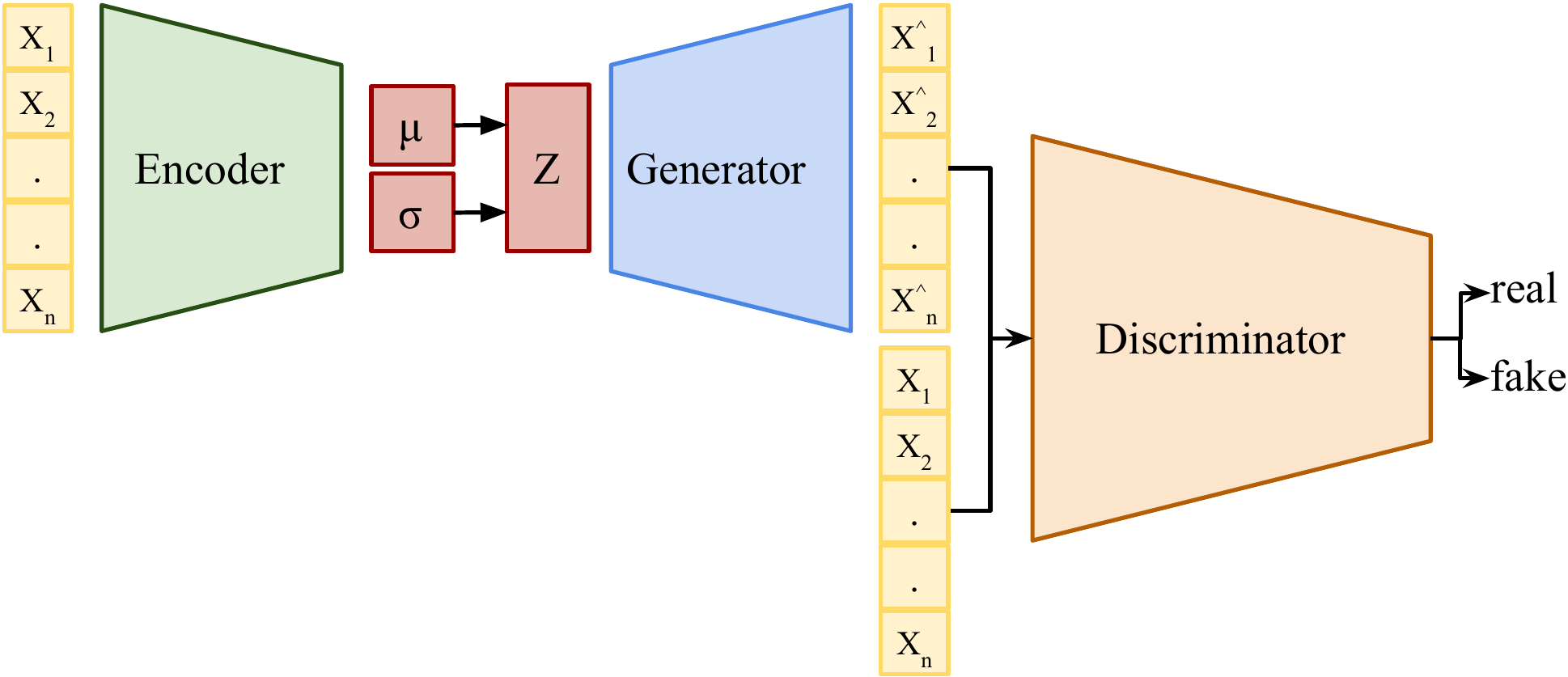}
        \caption{\small VAE-GAN model architecture. In this network, the encoder module maps the input sequence to the mean and the variance of a latent space with Gaussian distribution. The generator module reconstructs the input sequence from the latent space and tries to mislead the discriminator module to discriminate the generated sequence as a real sample. The discriminator module learns the distribution difference in real and fake samples 
        %\notered{how about we make this double column and fig2 and 3 single ? You can try and decide which one looks best.}
        }
        \label{fig:model}
\end{figure}

\begin{figure*}[htbp]
    \centering
    \captionsetup{justification=centering}
    \includegraphics[width=0.68\linewidth]{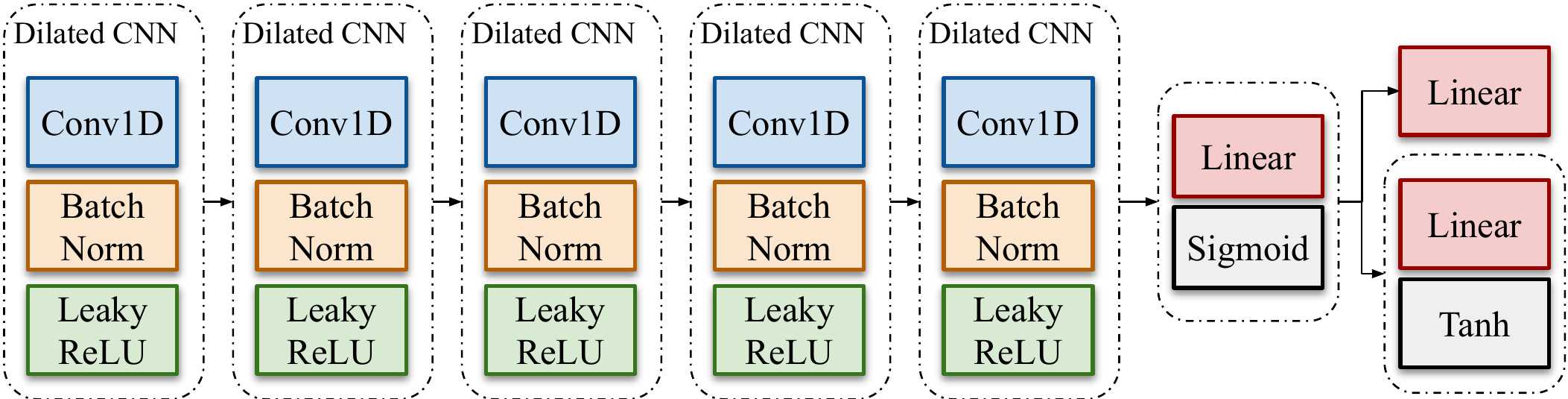}
    \caption{\small VAE-GAN \textbf{encoder} module structure}
    \label{fig:VAE-enc}
\end{figure*}
\begin{figure*}[htbp]
    \centering
    \captionsetup{justification=centering}
    \includegraphics[width=0.6\linewidth]{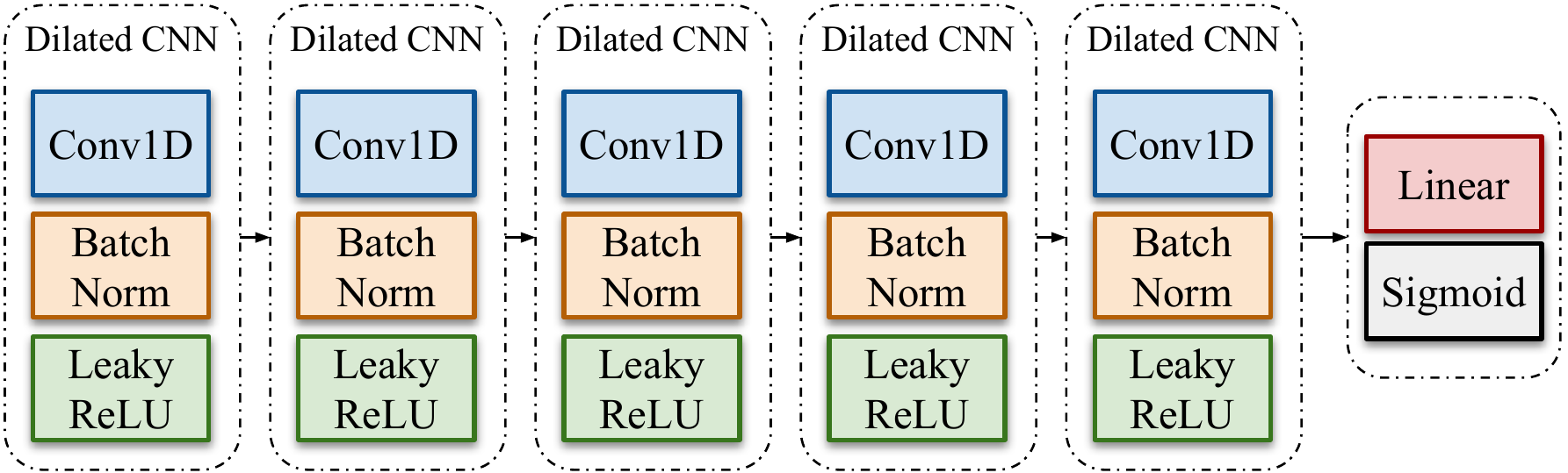}
    \caption{\small VAE-GAN \textbf{generator} and \textbf{discriminator} modules structure.}
    \label{fig:VAE-gen}
\end{figure*}

\subsection{Variational Autoencoder (VAE)}
\label{VAEdescription}
Autoencoder neural networks consist of two deep-learning based modules: encoder and decoder. The encoder module maps the input sequence into a meaningful latent space based on the original input sequence distribution, allowing the decoder module to reconstruct the input sequence with minimal error. However, vanilla autoencoders suffer from a lack of regularity in the latent space, which means the latent space may not be continuous to interpolate for data points that are not present in the input sequence. 

Variational autoencoders overcome this shortcoming by adding a regularization parameter, Kullback–Leibler (KL) divergence (equation(\ref{eq:kldiv})), in the training process, to ensure the latent space follows a Gaussian distribution. Instead of mapping the input sequence ($x$) to a vector, the VAE encoder ($E$) maps the data to two different vectors that are mean and standard deviation parameters of a Gaussian distribution. By minimizing the $L_{prior}$ loss, the encoder network is forced to compress the input sequence into a Gaussian distribution. In addition, it helps the decoder with reconstruction robustness, since the decoder module samples from a continuous distribution. The decoder loss is computed based on the distance between the reconstructed sequence ($\hat{x}$) and $x$. $L_{prior}$ and $L_{reconstruction}$ are backpropagated through the network to train the VAE parameters. 
\begin{dmath}
    L_{prior} = D_{KL}(E(x) \vert\vert \mathcal{N}(0,1))
    \label{eq:Lprior}
\end{dmath}
\begin{dmath}
    L_{reconstruction} = L_{prior} + \lVert \hat{x}-x \rVert^{2}
    \label{eq:Lrecon}
\end{dmath}

\subsection{Data-driven Generative Model}
The VAE-GAN architecture that is used for smart home synthetic data generation is shown in Fig.~\ref{fig:model}. This network architecture includes a GAN network with the generator module being a VAE neural network. As previously stated, the vanilla GAN network suffers from mode collapse. The main reason is that the discriminator is trapped in a local minimum, and the generator module repeatedly produces the output that is most likely to mislead the discriminator. As a result, training the GAN network becomes challenging and problematic. To address this problem, Larsen et al.~\cite{pmlr-v48-larsen16} inserted a variational autoencoder into the GAN's generator module to leverage the VAE latent space's regularity. %\notered{this sounds like we are the ones who invent this. cite the original paper here.}

The \textbf{encoder} module ($E$) compresses the input sequence into two vectors that are $mean_{z}$ and $variance_{z}$ of a Gaussian distribution by minimizing $L_{prior}$.

The \textbf{generator} module ($G$) reconstructs the input sequence from the latent space $z$ so that the reconstructed and original sequences have the lowest Mean Squared Error (MSE) by minimizing $L_{reconstruction}$. In addition, the input sequence cannot be considered as a generated sequence by the discriminator module, hence $L_{dG}$ must be kept to a minimum. $L_{Generator}$ is computed as in equation~(\ref{eq:Lgen}).

The \textbf{discriminator} module ($D$) needs to distinguish the original input sequence ($L_{Real}$) from the generator output sequence ($L_{fake}$). Meanwhile, to prevent the discriminator from failing to converge, $L_{noise}$ is added to the discriminator's loss function. This term enforces $D$ to distinguish a random sample from normal distribution from the real input sequence. The overall discriminator loss function is computed based on equation~(\ref{eq:LD}).
\begin{dmath}
    L_{dG} = E_{x}[log(D(G(z)))] 
    \label{eq:Ldgen}
\end{dmath}
\begin{dmath}
    L_{Generator} = L_{reconstruction} + L_{dG} 
    \label{eq:Lgen}
\end{dmath}
\begin{dmath}
    L_{real} = E_{x}[log(D(x))]
    \label{eq:Lreal}
\end{dmath}
\begin{dmath}
    L_{fake} = E_{z}[1-log(D(G(z)))]
    \label{eq:Lfake}
\end{dmath}
\begin{dmath}
    L_{noise} = E_{z}[1-log(D(\mathcal{N}(0,1)))
    \label{eq:Lnoise}
\end{dmath}
\begin{dmath}
    L_{D} = L_{real} + L_{fake} + L_{noise} 
    \label{eq:LD}
\end{dmath}

Fig.~\ref{fig:VAE-enc} presents the encoder structure of VAE-GAN. The generator and discriminator modules have similar structures, as shown in Fig.~\ref{fig:VAE-gen}. Dilated one-dimensional convolutional (\textbf{Dilated CONV1D}) neural network is used in the structure of the encoder, generator, and discriminator of the VAE-GAN network. This architecture is inspired by the WaveNet network~\cite{oord2016wavenet} and utilizes dilated causal convolution layers to capture long-term dependencies in the input sequence. The 1-dimensional convolution slides a filter on an input series by one stride. However, in the dilated convolution, the sliding filter skips the input sequence with certain steps while keeping the order of the input data. %Thus, the model will not learn from future data. 
Furthermore, multiple stacked dilated convolutional layers allow for longer input sequences, which reduces network complexity and training time compared with other long-term learning neural networks.

% \begin{figure}[htbp]
% \begin{minipage}[t]{0.52\linewidth}
%     \captionsetup{justification=centering}
%     \includegraphics[width=\linewidth]{images/VAE-GAN-Encoder.pdf}
%     \caption{\small VAE-GAN Encoder module structure \noteblue{Hi Mina, is it possible to use the former figures, which is a longer one? It seems that one is easier to be understood? What do you think.}}
%     \label{fig:VAE-enc}
% \end{minipage}%
%     \hfill%
% \begin{minipage}[t]{0.38\linewidth}
%     \captionsetup{justification=centering}
%     \includegraphics[width=\linewidth]{images/VAE-GAN-Generator.pdf}
%     \caption{\small VAE-GAN Generator and Discriminator modules structure}
%     \label{fig:VAE-gen}
% \end{minipage}
% \end{figure}

%% file: 4_Experiments.tex
\section{Evaluation Study}
\label{Experiments}

\subsection{Experiment Setup}
\label{Experimentsetup}
We use a real-world dataset, the iHomeLab RAPT dataset, to conduct our research \cite{huber2020residential}. This dataset includes residential electrical consumption data in appliance-level and aggregated household-level and solar panel (PV) energy production for five households in Switzerland spanning a period of 1.5 to 3.5 years with 5 minutes sampling frequency. 

The residential house we selected from the dataset has 594 days worth of training after data cleansing. We use aggregated energy consumption and PV power production data with a 15-minute resolution in our training process. A household's historical energy consumption and PV generation data are both time-series data. We feed the data to the network in a sequence of 96 consecutive points to preserve the data characteristics in temporal order (a whole day).

In addition to comparisons of our method with the real data distributions, we used the vanilla GAN neural network, which is another well-known deep learning-based generative model, as a baseline to evaluate the performance of the proposed synthetic smart home data generator model.

\subsection{Performance Metrics}
\label{Experimentindices}
In this section, we introduce several metrics to evaluate the data generation performance of our technique.

\subsubsection{Kullback–Leibler (KL) divergence}
\label{kldivergence} The KL divergence metric is used to determine the matching distance between two probability distributions. Equation (\ref{eq:kldiv}) shows the definition of KL divergence, where $p(x_{i})$ and $q(y_{i})$ are probability distributions. $D_{KL}(p \lVert q)=0$ represents two perfectly matched probability distributions that have
identical quantities of information, and $D_{KL}(p \lVert q)=1$ represents two completely different probability distributions.
\begin{equation}
    D_{KL}(p \lVert q) =\sum_{i=1}^{N}p(x_{i})log(\frac{p(x_{i})}{q(y_{i})})
    \label{eq:kldiv}
\end{equation}

\subsubsection{Maximum Mean Discrepancy (MMD)}
\label{MMD} The MMD metric applies a kernel to determine the distances between two distributions based on the similarity of their moments. We use the radial basis function (RBF) defined as equation (\ref{eq:rbf}) as the kernel in our experiments. Given the distributions $p(x)$ for $\{x_i\}_{i=0}^{N}$ and $q(y)$ for $\{y_j\}_{j=0}^{M}$, the MMD measure is calculated according to:
\begin{dmath}
    MMD(p,q)^2 = \frac{1}{N^2} \sum_{i=1}^{N} \sum_{j=1}^{N} K(x_{i},x_{j})- \frac{2}{MN} \sum_{i=1}^{N} \sum_{j=1}^{M} K(x_{i},y_{j})+ \frac{1}{M^2} \sum_{i=1}^{M} \sum_{j=1}^{M} K(y_{i},y_{j})
    \label{eq:mmd}
\end{dmath}
\begin{dmath}
    K(x,y)=\exp(\frac{-{\lVert x-y \rVert}^2}{2\sigma^2})
    \label{eq:rbf}
\end{dmath}

\subsubsection{Wasserstein Distance}
\label{Wdistance} The Wasserstein distance between the distributions $p(x)$, $q(y)$ is computed by:
\begin{dmath}
    l_{1}(p,q) = \underset{\pi \in \Gamma(p,q)}{inf} \int_{\mathbb{R} \times \mathbb{R}} \vert x-y \vert d \pi(x,y)
    \label{eq:wass}
\end{dmath}

Here $\Gamma(p,q)$ is the set of all pairs of random variables $\pi (x,y)$ with respective cumulative distributions $p$ and $q$. This metric computes the amount of distribution weight multiplied by the distance it has to be moved to transform distribution $p$ to $q$.

\subsubsection{Statistical parameters}
\label{stats} \cite{WANG2020110299} quantifies the load shape of a building's daily electricity consumption using five essential parameters: 
\begin{itemize}
    \item Near-peak load: $p_{peak}$ is any load value that exceeds $97.5$ percent of the load measurements.
    \item Near-base load: $p_{base}$ is the $2.5th$ percentile of daily load.
    \item High-load duration: duration of having constant load with values close to the near-peak load.
    \item Rise time: the amount of time needed from reaching near-peak load from near-base.
    \item Fall time: the amount of time needed to fall from near-peak load to near-base load.
\end{itemize}
We compute the mean and standard deviation of these essential parameters for each generative model along with the real-world electrical load and PV production data for better comparison.

\subsection{Evaluation Results}
\label{results}
Table~\ref{tab:metricresults} summarizes the KL divergence, Wasserstein distance, and MMD results between the real-world data and the synthetic data generated by GAN, and VAE-GAN networks. %\notered{sort them in the same order as you describe them above or change the order above since the rest of the text is following the table's order.} 

\begin{figure*}
    \centering
    \captionsetup{justification=centering}
    \begin{subfigure}{0.45\textwidth}
        \includegraphics[width=\linewidth]{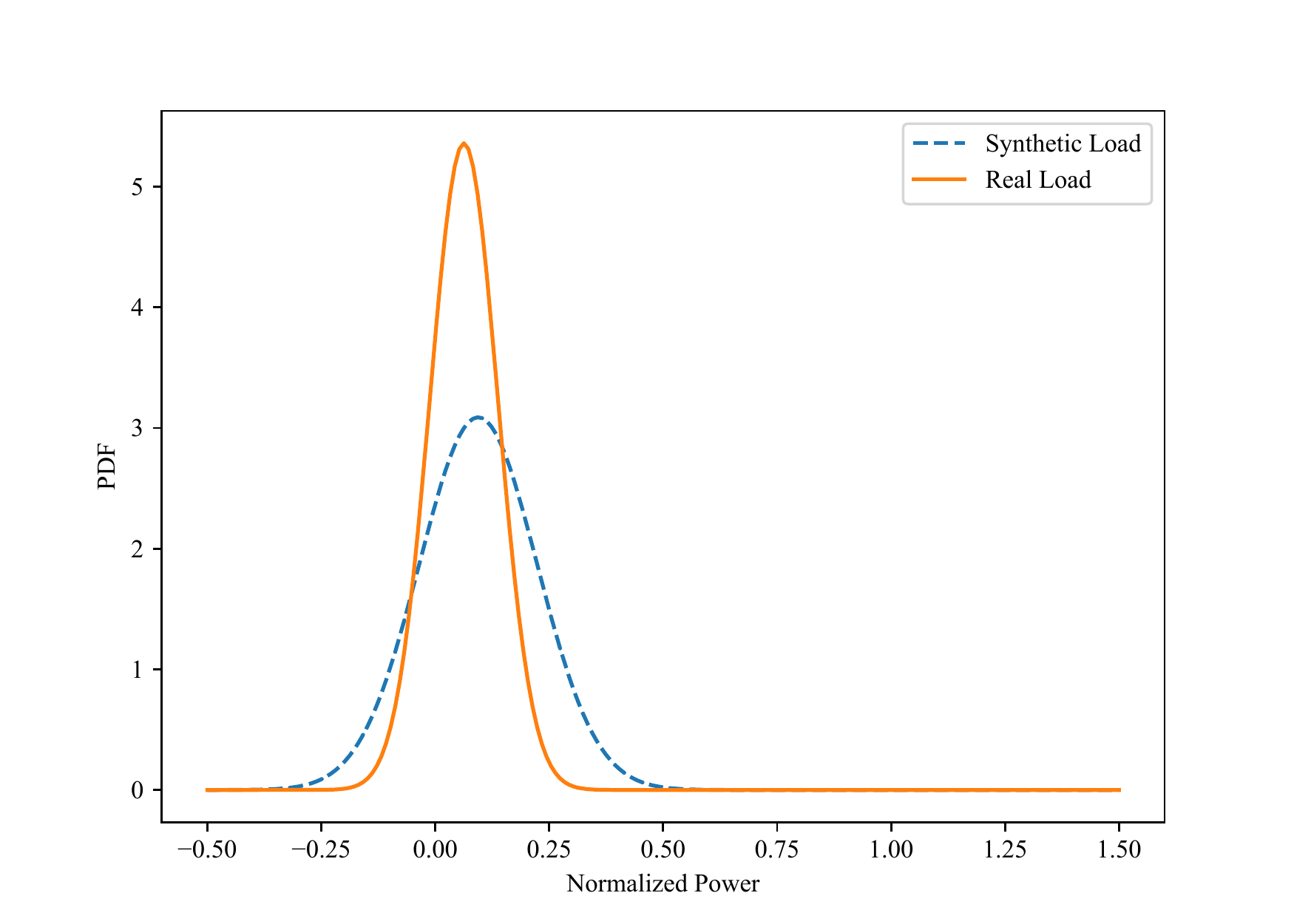}
        \subcaption{GAN}
        \label{fig:ganloaddist}
    \end{subfigure}%
    \hfill%
    \begin{subfigure}{0.45\textwidth}
        \includegraphics[width=\linewidth]{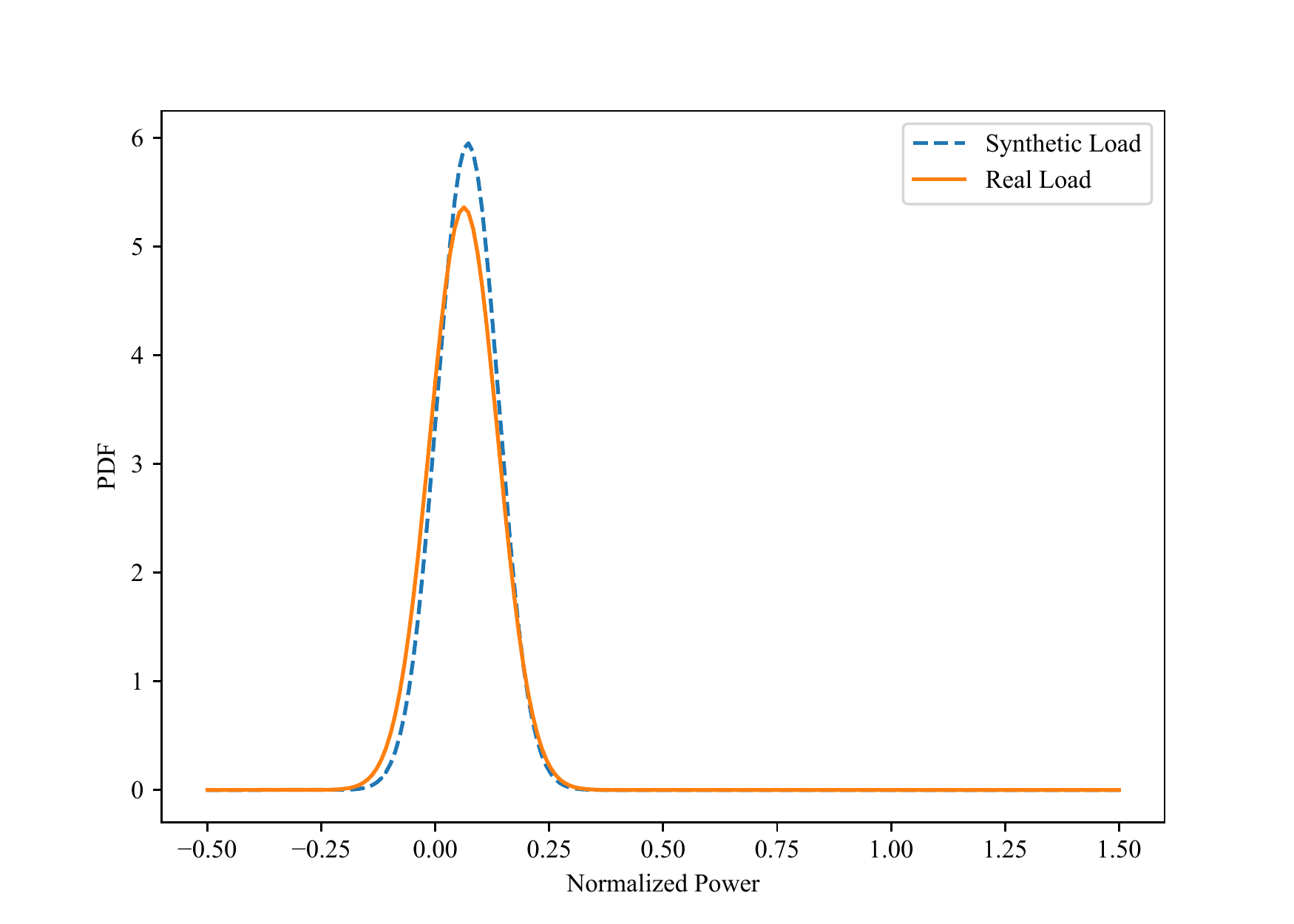}
        \subcaption{VAE-GAN}
        \label{fig:vaeganloaddist}
    \end{subfigure}%
    \caption[]{\small Electrical load consumption real and synthetic data probability density function for (a) GAN, and (b) VAE-GAN generative models. The blue line shows the real data PDF, the orange line shows the synthetic data PDF.}
    \label{fig:loaddists}
\end{figure*}

\begin{figure*}
    \centering
    \captionsetup{justification=centering}
    \begin{subfigure}{0.45\textwidth}
        \includegraphics[width=\linewidth]{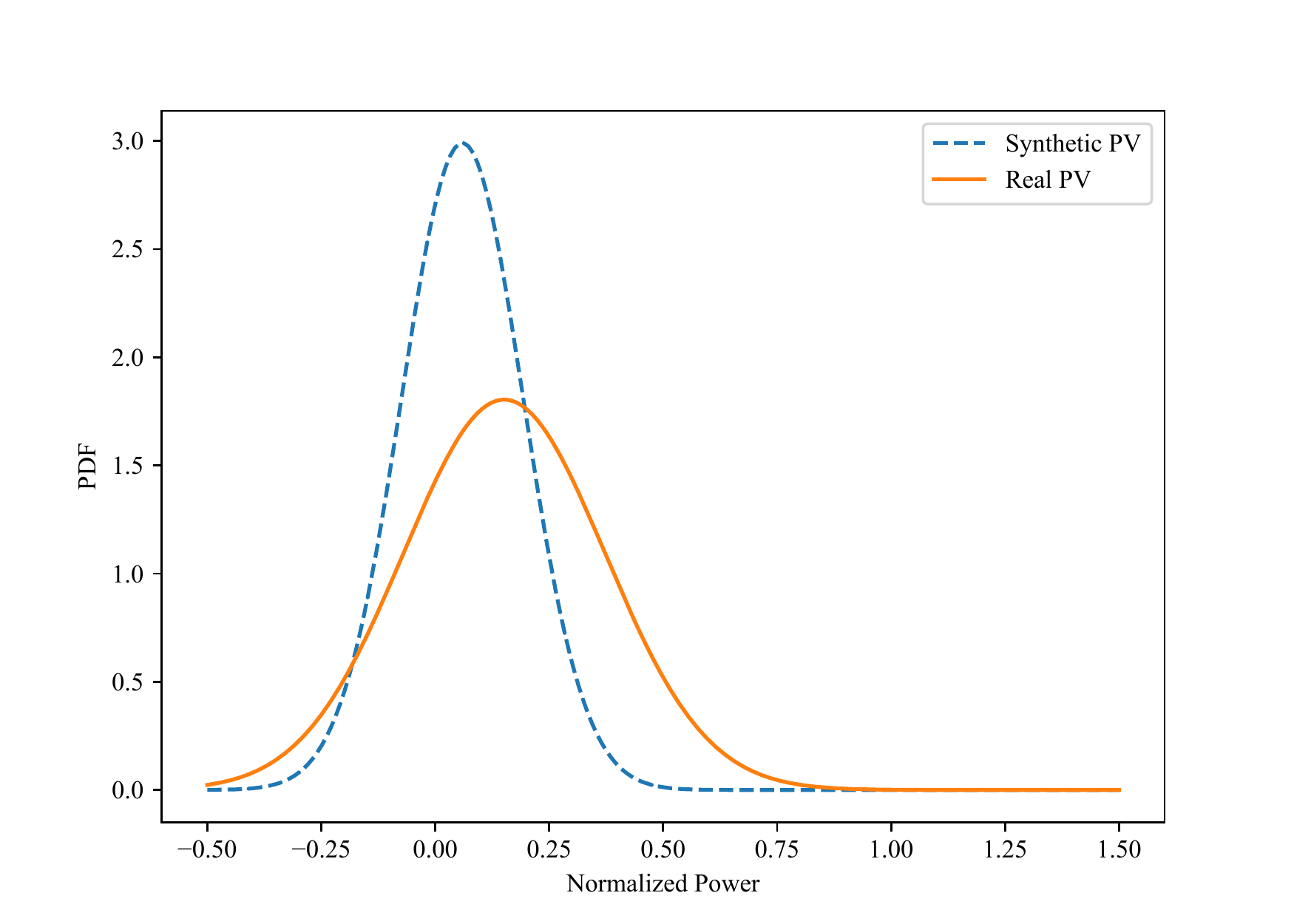}
        \subcaption{GAN}
        \label{fig:ganpvdist}
    \end{subfigure}%
    \hfill%
    \begin{subfigure}{0.45\textwidth}
        \includegraphics[width=\linewidth]{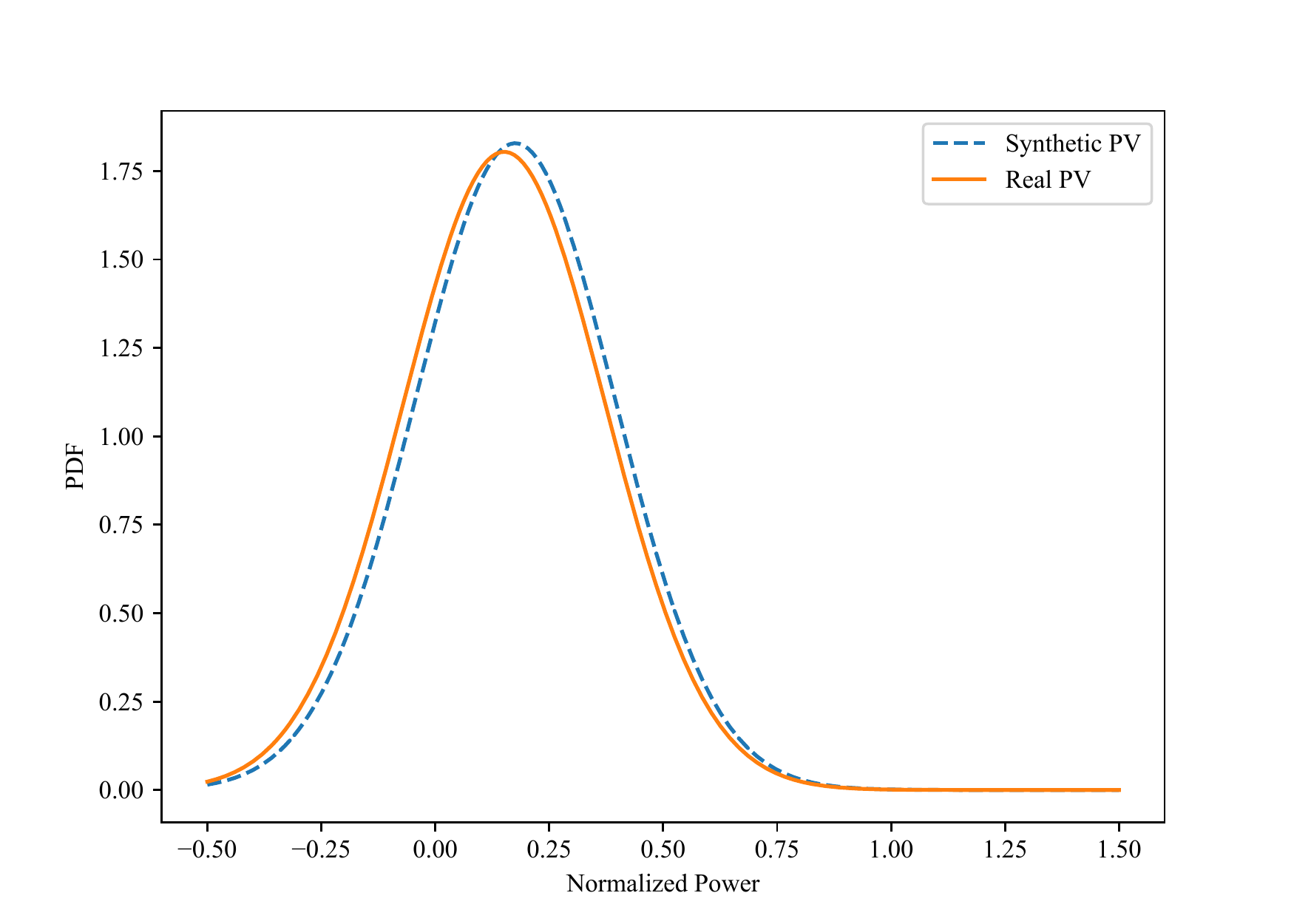}
        \subcaption{VAE-GAN}
        \label{fig:vaeganpvdist}
    \end{subfigure}%
    \caption[]{\small PV power production real and synthetic data probability density function for (a) GAN, and (b) VAE-GAN generative models. The blue line shows the real data PDF, the orange line shows the synthetic data PDF.}
    \label{fig:pvdists}
\end{figure*}

The synthetic data generated by the VAE-GAN network  (PV production and load consumption) have almost the same distribution as the real data, according to KL divergence results. As the KL divergence inclines more toward zero, two probability distributions are more similar. The probability density functions illustrated in Fig.~\ref{fig:loaddists} for electrical load consumption synthetic and real data, and Fig.~\ref{fig:pvdists} for PV power production synthetic and real data, supports the same claim. This demonstrates that the VAE-GAN network is capable of learning the smart home data distribution and producing plausible samples that reflect the same distribution. PV production data has a lower KL divergence than load consumption data for both networks, which is expected given that PV production data follows the sunrise and sunset pattern. It is still highly affected by environmental factors but has more seasonality than electrical load consumption data.

Wasserstein distance between synthetic data generated by the VAE-GAN network and real-world data is about 52\% and 72\% less compared to the data generated by the GAN network for electrical load consumption and PV production, respectively. This metric reveals that, compared with the GAN network, the distribution of synthetic data generated by the VAE-GAN network is considerably closer to the true distribution. Wasserstein distance values are large since the smart home data spectrum is between 0 and 15 $kW$.

MMD results, in Table~\ref{tab:metricresults}, show the difference between synthetic data and real data distributions are lower for the VAE-GAN network,   demonstrating the network's superior performance in generating synthetic data distributions compared to the GAN network.

Table~\ref{tab:statresults-load} and \ref{tab:statresults-pv} summarize the mean and standard deviation of five essential statistical parameters for synthetic and real electricity consumption and PV production data, respectively. For each parameter, we bold the result that is closer to the true distribution, for convenience. The data is min-max normalized before the training phase, then the base load is extremely close to zero. This is because of the difference in base load mean and standard deviation for synthetic and actual data for aggregated load consumption. The VAE-GAN network's ability to learn the load consumption and PV production data pattern is demonstrated by better peak load and high load duration results. Both models perform similarly and satisfactorily for the rise and fall time parameters.

\begin{table}
\vspace{2mm}
\captionsetup{justification=centering}
\caption{Distance between real and synthetic smart grid data distribution.}
\label{tab:metricresults}
% \resizebox{\linewidth}{!}{%
\noindent\begin{tabularx}{\linewidth}{@{}l*{7}{C}c@{}}
\toprule
\multicolumn{1}{c}{\multirow{2}{*}{\textbf{Model}}} &
\multicolumn{2}{c}{KL divergence} & \multicolumn{2}{c}{Wasserstein distance} & \multicolumn{2}{c}{MMD} \\ \cmidrule(l){2-3} \cmidrule(l){4-5} \cmidrule(l){6-7}
\multicolumn{1}{c}{} & \multicolumn{1}{c}{Load} & \multicolumn{1}{c}{PV} &
\multicolumn{1}{c}{Load} & \multicolumn{1}{c}{PV} &
\multicolumn{1}{c}{Load} & \multicolumn{1}{c}{PV}\\ \midrule
GAN  & 0.543 & 0.273  &530.3& 961.5& 0.227 & 0.224\\
\addlinespace
VAE-GAN & 0.017 & 0.006 &255.3 &266.6 & 0.101 & 0.117\\
\bottomrule
\end{tabularx}%
% }
\end{table}

% \lipsum[2-4]

\begin{table*}
\caption{Aggregated Load Consumption Evaluation Results. The numbers that are closest to the real data are highlighted in bold.}
\label{tab:statresults-load}
\resizebox{\textwidth}{!}{%
\noindent\begin{tabularx}{\textwidth}{@{}l*{11}{C}c@{}}
\toprule
\multicolumn{1}{c}{\multirow{2}{*}{\textbf{Model}}} &
\multicolumn{2}{c}{Base Load} & \multicolumn{2}{c}{Peak Load} & \multicolumn{2}{c}{High-Load Duration} & \multicolumn{2}{c}{Rise Time} & \multicolumn{2}{c}{Fall Time} \\ \cmidrule(l){2-3} \cmidrule(l){4-5} \cmidrule(l){6-7} \cmidrule(l){8-9} \cmidrule(l){10-11}
\multicolumn{1}{c}{} & \multicolumn{1}{c}{mean} & \multicolumn{1}{c}{std} &
\multicolumn{1}{c}{mean} & \multicolumn{1}{c}{std} &
\multicolumn{1}{c}{mean} & \multicolumn{1}{c}{std} &
\multicolumn{1}{c}{mean} & \multicolumn{1}{c}{std} &
\multicolumn{1}{c}{mean} & \multicolumn{1}{c}{std}\\ \midrule
%VAE & 22.75 & 141.99 & 23.56 & 147.01 & 0.02 & 0.93 & 0.48 & 0.84 & 0.47 & 0.81\\
%\addlinespace
GAN  & \textbf{3.53} & \textbf{22.34} & 257.84 & 1658.12 & \textbf{0.01} & 0.08 & \textbf{0.45} & \textbf{0.80} & \textbf{0.49} & \textbf{0.96}\\
\addlinespace
VAE-GAN  & 1.73 & 10.99 & \textbf{119.85} & \textbf{752.92} & \textbf{0.00} & 0.04 & \textbf{0.44} & 0.74 & 0.52 & \textbf{0.97} \\
\addlinespace
Real data & 9.75 & 51.13 & 151.39 & 1008.20 & 0.02 & 0.33 & 0.48 & 0.87 & 0.49 & 0.91 \\
\bottomrule
\end{tabularx}%
}
\end{table*}

\begin{table*}
\caption{PV Power Production Evaluation Results. The numbers that are closest to the real data are highlighted in bold.}
\label{tab:statresults-pv}
\resizebox{\textwidth}{!}{%
\noindent\begin{tabularx}{\textwidth}{@{}l*{11}{C}c@{}}
\toprule
\multicolumn{1}{c}{\multirow{2}{*}{\textbf{Model}}} &
\multicolumn{2}{c}{Base Load} & \multicolumn{2}{c}{Peak Load} & \multicolumn{2}{c}{High-Load Duration} & \multicolumn{2}{c}{Rise Time} & \multicolumn{2}{c}{Fall Time} \\ \cmidrule(l){2-3} \cmidrule(l){4-5} \cmidrule(l){6-7} \cmidrule(l){8-9} \cmidrule(l){10-11}
\multicolumn{1}{c}{} & \multicolumn{1}{c}{mean} & \multicolumn{1}{c}{std} &
\multicolumn{1}{c}{mean} & \multicolumn{1}{c}{std} &
\multicolumn{1}{c}{mean} & \multicolumn{1}{c}{std} &
\multicolumn{1}{c}{mean} & \multicolumn{1}{c}{std} &
\multicolumn{1}{c}{mean} & \multicolumn{1}{c}{std}\\ \midrule
%VAE & 1.33 & 8.36 & 5.34 & 33.35 & 0.01 & 0.19 & 0.51 & 0.88 & 0.45 & 0.76\\
%\addlinespace
GAN  &\textbf{ 0.00} & \textbf{0.00} & 148.53 & 927.53 & \textbf{0.02} & \textbf{0.50} & \textbf{0.49} & 0.91 & 0.45 & 0.81 \\
\addlinespace
VAE-GAN  & \textbf{0.04} & 0.29 & \textbf{205.63} & \textbf{1283.73} & \textbf{0.00} & 0.00 & 0.42 & 0.79 & \textbf{0.54} & \textbf{1.02}\\
\addlinespace
Real data & 0.00 & 0.00 & 197.91 & 1236.58 & 0.01 & 0.35 & 0.73 & 5.46 & 0.68 & 4.80 \\
\bottomrule
\end{tabularx}%
}
\end{table*}

%% file: 5_Conclusion.tex
\section{Conclusion}
\label{Conclusion}
Synthetic data generation is an important capability to apply advanced data science and machine learning techniques for smart home management. In this work, we apply a variational autoencoder GAN (VAE-GAN) for synthetic time series generation of the smart home data including electrical load and PV generation. The main advantage of this technique is that the network does not require any prior training analysis on the data, and can be utilized to generate different forms of smart home data, including household electricity load and PV power generation. The model performance and synthetic data effectiveness are assessed with respect to the vanilla GAN network. The simulations show that our proposed scheme achieves satisfying performance. 
%\notered{References have some mistakes please fix. for example 1 and 2 are the same.}  

%% file: 6_Acknowledgement.tex
\section*{Acknowledgement}
Hao Zhou and Melike Erol-Kantarci were supported by the Natural Sciences and Engineering Research Council of Canada (NSERC), Collaborative Research and Training Experience Program (CREATE) under Grant 497981 and Canada Research Chairs Program.